\documentclass[conference]{IEEEtran}
\IEEEoverridecommandlockouts
\usepackage{cite}
\usepackage{amsmath,amssymb,amsfonts}
\usepackage{algorithmic}
\usepackage{graphicx}
\usepackage{textcomp}
\usepackage{xcolor}
\def\BibTeX{{\rm B\kern-.05em{\sc i\kern-.025em b}\kern-.08em
    T\kern-.1667em\lower.7ex\hbox{E}\kern-.125emX}}
\begin{document}

\renewcommand{\eqref}[1]{eq.~\ref{#1}}
\newcommand{\Eqref}[1]{Eq.~\ref{#1}}
\newcommand{\figref}[1]{Figure~\ref{#1}}

\newcommand{\secref}[1]{\S \ref{#1}}
\newcommand{\appref}[1]{appendix \ref{#1}}
\newcommand{\Appref}[1]{Appendix \ref{#1}}

\title{Embodied vision for learning object representations }

\author{\IEEEauthorblockN{Arthur Aubret}
\IEEEauthorblockA{
\textit{University Clermont Auvergne} \\ \textit{CNRS, Pascal institute}\\
Clermont-Ferrand, France \\
arthur.aubret@uca.fr}
\and
\IEEEauthorblockN{Céline Teulière}
\IEEEauthorblockA{\textit{University Clermont Auvergne} \\ \textit{CNRS, Pascal institute} \\
Clermont-Ferrand, France \\
celine.teuliere@uca.fr}
\and
\IEEEauthorblockN{Jochen Triesch}
\IEEEauthorblockA{\textit{Frankfurt Institute for Advanced Studies} \\
Frankfurt am Main, Germany \\
triesch@fias.uni-frankfurt.de}
}

\maketitle

\begin{abstract}
Recent time-contrastive learning approaches manage to learn invariant object representations without supervision. This is achieved by mapping successive views of an object onto close-by internal representations. When considering this learning approach as a model of the development of human object recognition, it is important to consider what visual input a toddler would typically observe while interacting with objects. First, human vision is highly foveated, with high resolution only available in the central region of the field of view. Second, objects may be seen against a blurry background due to infants' limited depth of field. Third, during object manipulation a toddler mostly observes close objects filling a large part of the field of view due to their rather short arms.  Here, we study how these effects impact the quality of visual representations learnt through time-contrastive learning. To this end, we let a visually embodied agent ``play'' with objects in different locations of a near photo-realistic flat. During each play session the agent views an object in multiple orientations before turning its body to view another object. The resulting sequence of views feeds a time-contrastive learning algorithm. Our results show that visual statistics mimicking those of a toddler improve object recognition accuracy in both familiar and novel environments. We argue that this effect is caused by the reduction of features extracted in the background, a neural network bias for large features in the image and a greater similarity between novel and familiar background regions. 
We conclude that the embodied nature of visual learning may be crucial for understanding the development of human object perception.
\end{abstract}


\section{Introduction}

Babies learn to build representations of their surroundings before knowing the word for an object, suggesting that they perform a form of unsupervised learning \cite{quinn1993evidence}. What mechanism underlies their learning process? Several experiments show that modifying the temporal contiguity of an animal's visual experiences shapes its visual representations such that temporally close visuals inputs tend to be similarly represented in the visual cortex \cite{li2010unsupervised,wood2018development}. The overall idea that the brain uses temporal closeness to build visual representations without supervision has been called the \textit{slowness principle} \cite{wiskott2002slow}. One way to implement the slowness principle in machine learning is through \textit{contrastive learning}. While the original idea of contrastive learning was to learn visual representations (almost) invariant to data-augmentations like image crops or rotations, one can also frame data-augmentations as a shift in time \cite{schneider2021contrastive}; in fact, time-based augmentations like object rotations \cite{schneider2021contrastive} or saccades \cite{wang2021use} allow to build task-useful representations with performance close to less natural augmentations \cite{wang2021use}. Therefore, contrastive learning with time-based augmentations \cite{schneider2021contrastive} appears to be a strong candidate for understanding how animals construct invariant object representations.
In current machine learning research, contrastive learning methods mostly consider ImageNet-like datasets of high-resolution images with a great variety of differently-sized objects \cite{chen2020simple}. Yet, because of its embodiment, a toddler experiences very different kinds of visual inputs. First, the density of cones is higher in the fovea compared to the periphery, leading to poor resolution in peripheral vision \cite{strasburger2011peripheral}. Second, a child learns to control its distance of focus within a few months \cite{currie1997development}. Due to the limited depth of field, a properly focused object will be seen against a more or less blurry background. Third, the ability of a toddler to act onto its environment and the environment itself shape what the infant sees \cite{byrge2014developmental}. A typical infant experiences a relatively small set of rooms/outdoor environments \cite{sullivan2020saycam}, spending most of its time in the home or a daycare environment. Furthermore, 18-months toddlers have short arms, making objects they play with appear bigger in their field of view in comparison with adults \cite{bambach2018toddler,smith2011not}.




Here, we aim to study the impact of these visual statistics on representations learnt through the slowness principle. We take advantage of a near photo-realistic simulation platform, ThreeDWorld (TDW) \cite{gan2021threedworld} and a recent dataset of 3D models of toy objects \cite{stojanov2021using} to visually embody an agent like an 18-month-old toddler. We introduce a house environment in which we simulate this agent playing with different objects in different locations: we simulate object rotations, saccades, and switching between objects. These transformations induce natural time-based augmentations for training with time-contrastive learning. Our first experimental contribution demonstrates that the diversity/complexity of backgrounds can greatly hurt object recognition, thereby motivating the need for attenuating background information. In our second experimental contribution, we investigate the impact of four particular visual statistics on object recognition through time-contrastive learning: the foveation blurring the surroundings of a fixated object; the depth of field effect blurring objects in the background; the distance of the object from the agent; the range of saccades during object exploration. Our analysis is three-fold: 1) we show that an object-centered foveation and a restricted depth of field promote object representations that generalize to novel environments; 2) we exhibit that making small saccades attenuates the extraction of background information; 3) presumably due to a neural network bias, closer objects make easier object recognition, in line with previous results \cite{bambach2018toddler}. The combination of all three effects drastically improves object recognition accuracy, thereby showing that toddler-inspired visual inputs improve object representations.






\begin{figure}
    \centering
    \includegraphics[width=1\linewidth]{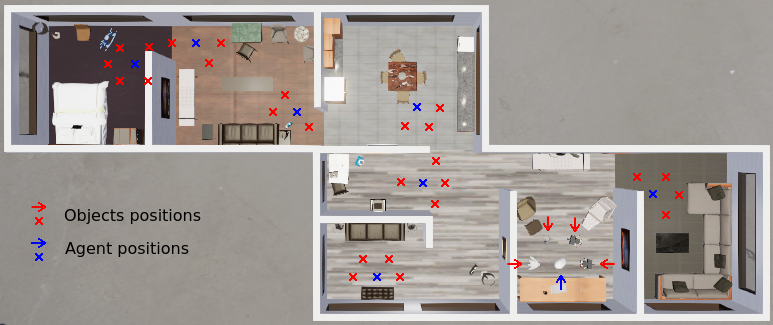}
    \caption{Scene of TDW used to situate our agent in a house. In blue, we indicate possible agent positions and we display in red possible object positions. In this episode, the agent is located in the office and can interact with four objects.}
    \label{fig:environments}
\end{figure}

\begin{figure}
    \centering
    \includegraphics[width=1\linewidth]{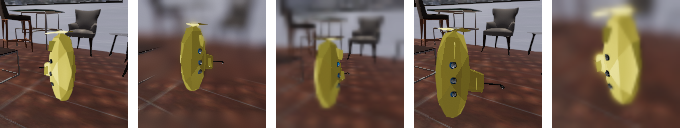}
    \caption{Submarine with different orientations, perceived through different eye orientation, but in front of a single background. From left to right, we illustrate: a low amplitude of saccades that keeps the object centered; the progressive blur of the scene according to its relative depth with the focused object (depth of field); the progressive blur as pixels get distant from the image center (foveation); the increase in size that follows the greatest closeness of the object; the combination of all transformations.}
    \label{fig:submarines}
\end{figure}




\section{Related works}



\paragraph{SimCLR-TT} SimCLR, a state-of-the-art unsupervised contrastive learning algorithm  \cite{chen2020simple}, recently managed to learn representations of visual images useful to downstream classification tasks. Intuitively, two semantically similar/dissimilar images should be respectively close/distant in the latent space, such that the distance between images in the latent space captures the semantics of the images. Following the slowness principle, SimCLR-TT \cite{schneider2021contrastive, oord2018representation} expects two semantically similar images to be successive in time. SimCLR-TT samples an image $x_i$ and selects its successive or previous image (in time) to be its positive sample $x_j$. Then, SimCLR-TT samples $N-1$ other images $x_k$ different from $x_i$, and their predecessor/successor as $x_{N+k}$. Thereafter, SimCLR-TT minimizes the loss function of SimCLR by adapting the parameters $\theta$ of the representation encoder (a neural network) $f_{\theta}$. Thus, it minimizes the following loss function for all positive pairs:

\begin{equation}
    \mathcal{L}_{i,j}= -\log \frac{e^{\text{sim}(f_{\theta}(x_i), f_{\theta}(x_j))/\tau}}{\sum_{k=1,k\neq i, k\neq j}^{2N} e^{\text{sim}(f_{\theta}(x_i), f_{\theta}(x_k))/\tau}}, \label{eq:simclr}
\end{equation}

where $sim$ and $\tau$ are respectively a similarity function and the temperature hyper-parameter. The numerator of \eqref{eq:simclr} makes sure that successive images are close in the latent space. To avoid the collapse of the representation into a single vector, the denominator makes sure that all representations are relatively distant.

\paragraph{Visual data-augmentations} Most contrastive learning approaches use unrealistic data augmentations like color distortion, crop, rotation, resize, blur \cite{chen2020simple,grill2020bootstrap}, Jigsaw \cite{misra2020self}, selection of color channels \cite{tian2020makes}. In practice, it has been argued that these augmentations are related to a subset of more realistic augmentations \cite{laflaquiere2021sensorimotor}: crop can relate to occlusion, resize to object depth motion, color changes to illumination changes etc.  Whether unrealistic augmentations and their realistic counterpart truly result in similar representations remains unclear. \cite{ryali2021learning} proposed as a data-augmentation to detect the salient object and switch its background with one from another object; the resulting improved accuracy validates that discarding background information may improve the robustness of object representations. Another body of work tries to learn embeddings of videos thanks to the temporal contiguity of frames. Then, positive pairs are different clips (temporal crops) of the same video  \cite{feichtenhofer2021large,qian2021spatiotemporal,wang2021long} or overlapping clips with different temporal windows \cite{recasens2021broaden}. We refer to \cite{jaiswal2021survey} for a review of contrastive learning methods along with the different data-augmentation methods. Unlike previous visual contrastive learning approaches, we visually embody our agent in a 3D environment that allows for both ego-centric and object-centric transformations. 

\paragraph{Time-contrastive learning for object representations}
closer to our work, we emphasize three works that study the representations learnt through realistic transformations. Using videos, some approaches proposed to take advantage of the temporal information to learn image embeddings \cite{knights2021temporally,orhan2020self}. But datasets may not simulate the depth of field/foveation and do not allow the study of the complexity of the background. Similarly to us, \cite{schneider2021contrastive} consider 3D object rotations as transformations, but with a uniform background and without ego-motions. Interestingly, they show that long-time fixations improve the representations in comparison with short-time fixations. \cite{wang2021use} showed that foveation based on magnification and saccades can efficiently replace crop augmentations. Unlike us, they do not study the impact of visual statistics on the learnt representation.

\paragraph{Toddler-inspired machine learning} several works already studied the impact of some visual statistics on machine learning methods with respect to different objectives, but not in the context of time-contrastive learning. \cite{bambach2018toddler} showed that supervised learning could benefit from learning with images of objects, sized and centered as experienced by toddlers; in fact, agents may even move to create un-occluded views of an object to improve recognition \cite{yang2019embodied}. Machine learning simulations also support that the increase of acuity in the first months of life provides a natural curriculum for learning. This can aid the learning of reflexes, like the acquisition of vergence behavior (alignment of the two eyes on the same point) \cite{priamikov2015role} or the detection of binocular disparity (difference between the two eye images) \cite{dominguez2003developmental}. This can also aid learning representations of faces robust to different resolutions \cite{jang2021convolutional,vogelsang2018potential}.








\section{Method}\label{sec:method}
We want to simulate the visual observations experienced by toddlers playing with objects in order to study the impact of visual statistics on representations learnt through SimCLR-TT \cite{schneider2021contrastive}. To simulate the visual environment of infants, we consider a house environment, as illustrated in \figref{fig:environments}, and manually define different play locations. These locations can have similar/different floors, walls and objects, thereby making non-obvious their recognition. In a location, the agent engages in 100-step long play sessions (also called episodes), which are initially built in a 3-steps process. Firstly, we place the agent in the center of a randomly sampled location. Secondly, we put randomly sampled objects around the agent (between two and six according to the location and surrounding objects). Because of their different positions, a different background gets associated to each of them. The objects are randomly sampled from a set of $20$ (unless stated otherwise) textured and untextured objects randomly extracted from the Toys-4k dataset. We expect untextured objects to make the task harder, since the model can not focus on the color regardless of the shape of the object. Finally, we randomly turn the agent in front of an object. For example, in \figref{fig:environments}, the agent runs a play session in the office of the house and can play with four objects (each seen against a different background). 

To simulate viewing sequences experienced by a child during a play session, we introduce three factors of variation that change across the time of a play session. At a low frequency (every $10$ timesteps), the agent selects a new object (the first one to the right of the current object) and rotates its body in the direction of the object. At a higher frequency (every timestep), it rotates the object by an azimuth angle uniformly sampled in $[0; 30]$ and executes a saccadic eye movement by sampling its position from a normal distribution $S \times \mathcal{N}([0,0],\mathbb{I})$ where $S$ is a hyper-parameter denoting the saccade amplitude. Note that, at $[0,0]$, the eyes are aligned with the object such that it appears centered in the field of view. We expect this difference of frequency to make the agent discard the information about high-frequency variations (saccades and object rotation) while keeping information about low-frequency variations \cite{schneider2021contrastive}, \textit{i.e.}, object identity, background identity and location identity. The highly frequent transformations have different properties that we expect to differently impact the learnt features: the object rotation minimally impacts the background and the ego-centric saccades impacts both the object and the background. 

We now introduce the three visual effects shaping the agent's visual input, namely fovation, distance/centering of objects, and depth of field.

\subsection{Foveation and saccades}

In humans, photo-receptors called cones are distributed on the retina of each eye to detect light under day-light conditions. These cones are present with high density in the fovea and low density in the periphery of the retina \cite{szel1996distribution} such that humans have high visual accuracy in the center of gaze and low accuracy in the periphery (\textbf{foveation}). To simulate this change of acuity, we use a previous model of humans' visual acuity \cite{jiang2015salicon,perry2002gaze}. It creates a pyramid of multi-resolution images and interpolates each pixel between the different resolutions according to the distance of each pixel from the focus position. 

To simulate different amplitudes of \textbf{saccades}, and thus different levels of overlap between the object and the high-accuracy area, we consider a range saccade amplitudes $S \in \{3 , 2 , 1\}$. For $S=3$, this results in a range of saccade angles (yaw and pitch axis) with significant density in $[-15^\circ,15^\circ]$ degrees from the object center. Low amplitudes of saccades result in a strong average overlap between the high-accuracy area and the object position, and inversely for high amplitudes of saccades. In practice, while our range of amplitudes is similar to adult data \cite{trukenbrod2019spatial}, human saccades usually follow a non-normal density distribution. However, we expect our simple distribution of saccades to be a good enough approximation to quantify the impact of doing object-relevant saccades versus object-irrelevant saccades on object representations.


\subsection{Depth of field}

The accommodation reflex of humans describes their ability to focus on an object of interest depending on its distance from the observer in order to obtain a sharp image of the object. An important part of the reflex is the increase of the curvature of each eye lens, which adapts the refractive effect of the lens on the light to the desired focus distance. For instance, if one starts to look at the sky right after reading a book, the lens will decrease its curvature to have the distance of focus matching the distance of the sky. Out of focus objects will get progressively blurred as they lie on a plane distant from the focus plane. The range of depths around the distance of focus observed with \textit{acceptably sharp focus} is called the depth of field. In practice, according to light levels or the depth of the object of focus, the iris increases/decreases the depth of field by respectively decreasing/increasing the size of the pupil. To keep it simple, we set the focus distance of the camera to correspond to the distance of the manipulated toy and we decrease the depth of field by decreasing the \textbf{aperture} number of the TDW platform \cite{gan2021threedworld}. Overall, this makes the objects appear in focus while blurring the background, as shown in \figref{fig:submarines}.


\subsection{Object position/distance}

Between one and two years, infants start to walk, manipulate and play with objects. This learning step considerably changes what a child sees; while it was previously seeing very cluttered scenes of relatively small objects, it starts to watch close and centered objects \cite{bambach2018toddler,smith2011not}. To simulate this change of object distance, we bring objects \textbf{closer} to the agent by a factor of $0.7$. \figref{fig:submarines} illustrates the size change that results from decreasing the distance between the object and the agent. We do not assume this distance to be very realistic, but a true hand-based manipulation of the object is out of our scope and our choice avoids physical collisions. However, this modification allows us to qualitatively assess the impact of modifying the agent-to-object distance.

\section{Experiments}

We aim to evaluate whether considering the visual statistics of toddlers improves object recognition in time-contrastive learning, considering the typical environment of a toddler. 

During training, the agent accumulates observations in a buffer of size $80,000$ and, between each timestep, it trains the encoder network with SimCLR-TT \cite{schneider2021contrastive}. The $128\times 128$ RGB images go through a succession of convolution layers with the following [channels, kernel size, stride, padding] structure: [64, 8, 4, 2], [128, 4, 2, 1], [256, 2, 2, 1], [256, 2, 2, 1]. These are followed by an average pooling layer and a linear layer ending with $20$ units (unless stated otherwise). After each convolution layer, we apply a non-linear ReLU and a dropout layer ($p=0.5$) to prevent over-fitting \cite{srivastava2014dropout}. Our preliminary results showed that such a shallow architecture combined with Euclidean distance, close to what is found in visual RL \cite{mnih2015human} but augmented with dropout and average pooling, works better than standard architectures used in contrastive learning \cite{chen2020simple}. We hypothesize that this stems from the use of less diverse inputs. We apply a weight decay of $1e-6$ and update weights with the AdamW optimizer \cite{loshchilov2018decoupled} and a learning rate of $5e-4$.

For evaluation purposes, we train linear classifiers on top of the learnt representations to assess their quality \cite{chen2020simple}. We either train them to recognize the objects or the backgrounds. We use two datasets of at least $1200$ images (depending on the number of backgrounds): 1) a validation dataset composed of images generated similarly to the training data; 2) a test dataset composed of similar object views in unseen backgrounds (from another house). We apply the visual effects (foveation, depth of field, etc.) on the training images and on the validation and test datasets. In all experiments, we compute the mean and standard deviation over five random seeds.

\subsection{Diverse complex backgrounds impede object learning} \label{sec:backgrounds}

\begin{figure}[t]
    \centering
    \includegraphics[width=1\linewidth]{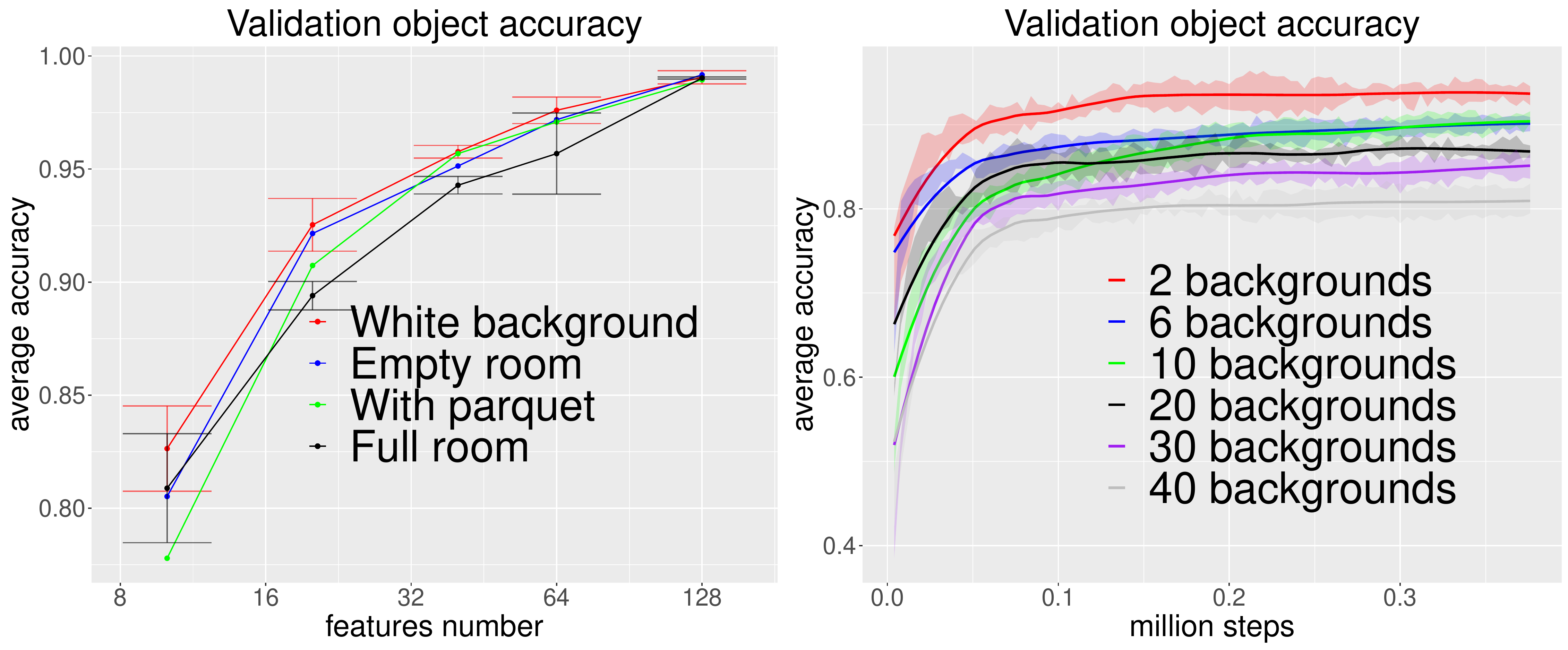}
    \caption{\textbf{Left:} Object recognition accuracy on the validation dataset as a function of dimensionality of the latent space for different levels of background complexity in the single-room environment, after $400k$ training steps/interactions. Error bars indicate the standard deviation of the \textit{white background} and \textit{full room} experiments, we omit those of other curves for clarity. \textbf{Right:} Object recognition accuracy on the validation dataset for different numbers of backgrounds seen by the agent. Shaded regions around the curves indicate one standard deviation. }
    \label{fig:singleroom}
\end{figure}

To study the effects of different backgrounds on object learning, we set up one location with six object positions/backgrounds and we progressively increase the diversity/complexity of features in the six backgrounds. Specifically, we train our representation with: fully white background almost without clues about what background is being watched; an untextured empty room with floor/wall corners apparent; the same empty room with an oriented parquet; a room with the same parquet, textured walls and several objects. The left part of \figref{fig:singleroom} suggests that increasing the complexity of the background has a minor effect on object representations, mostly for a medium size latent space (size of 20 and 40 and 64 neurons). 

\begin{figure}
    \centering
    \includegraphics[width=0.8\linewidth]{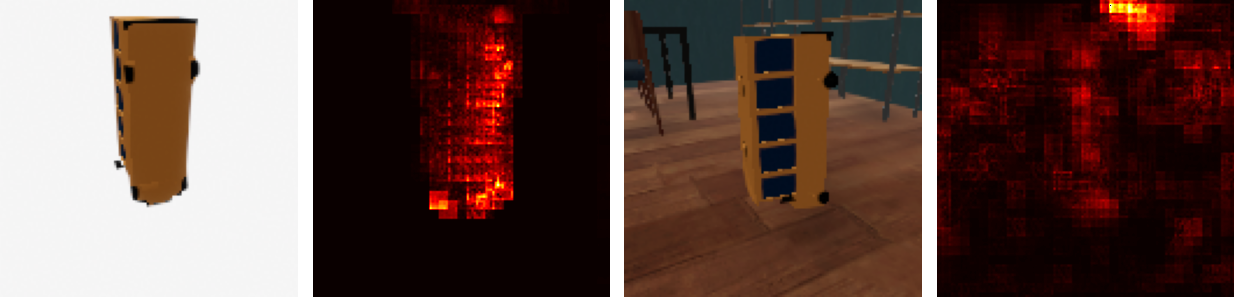}
    \caption{Saliency maps of the SimCLR loss on a bus toy seen against a white background (left) or a room with several background objects (right)}
    \label{fig:saliencies}
\end{figure}

To investigate the impact of the number of backgrounds on the representation, we set up different houses with increasingly higher number of different backgrounds (the sum of backgrounds for all possible locations). The right part of \figref{fig:singleroom} highlights that, as we increase the overall number of backgrounds, the object recognition accuracy decreases. To visualize what the encoder is extracting from the image depending on the number/complexity of the background, we display in \figref{fig:saliencies} typical saliency map \cite{simonyan2014deep} outputs by the encoder when applying the SimCLR loss, depending on the background of the image. We clearly observe that the encoder better focuses on the object when there is no distracting background. Overall, these experiments suggest that object recognition accuracy can be improved by reducing distracting information from the background. 

\subsection{Toddler-inspired visual input aids object learning}

\begin{figure}[t]
    \centering
    \includegraphics[width=1\linewidth, height=3.5cm]{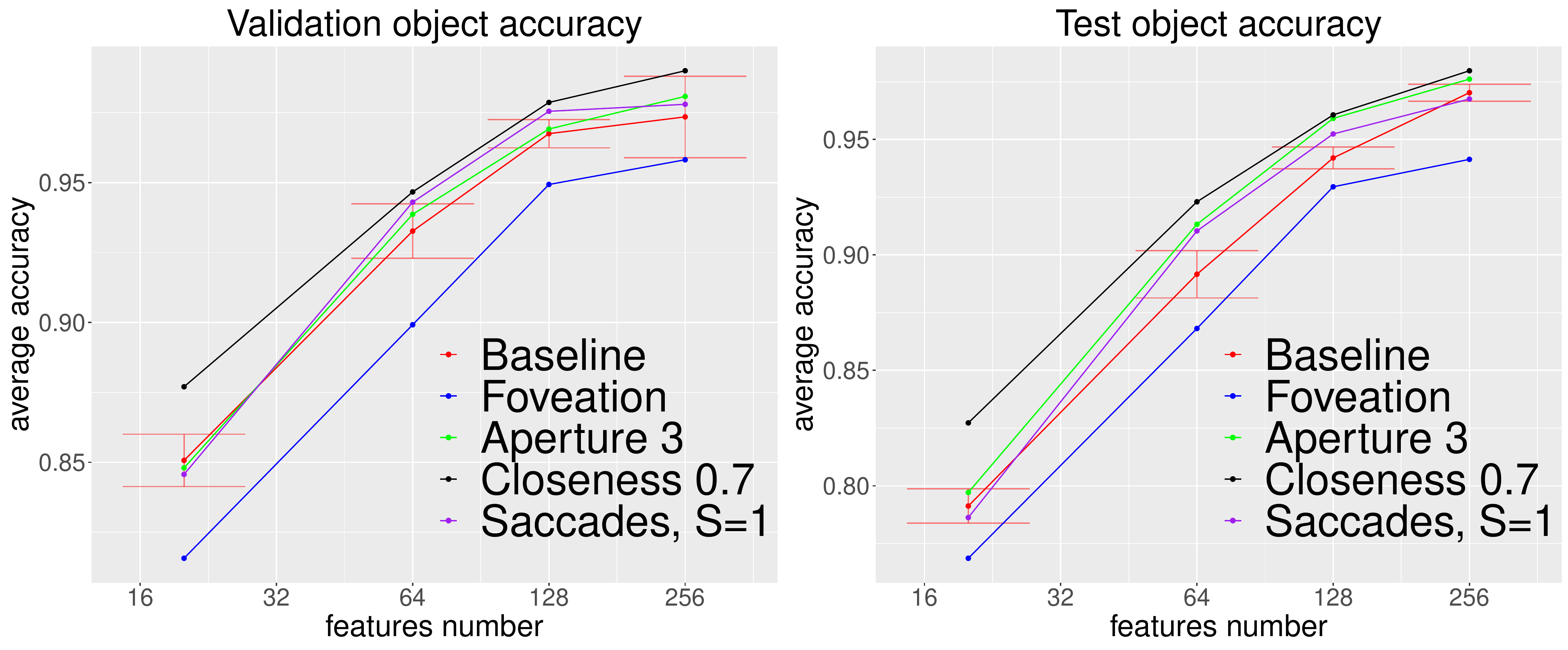}
    \includegraphics[width=1\linewidth,height=3.5cm]{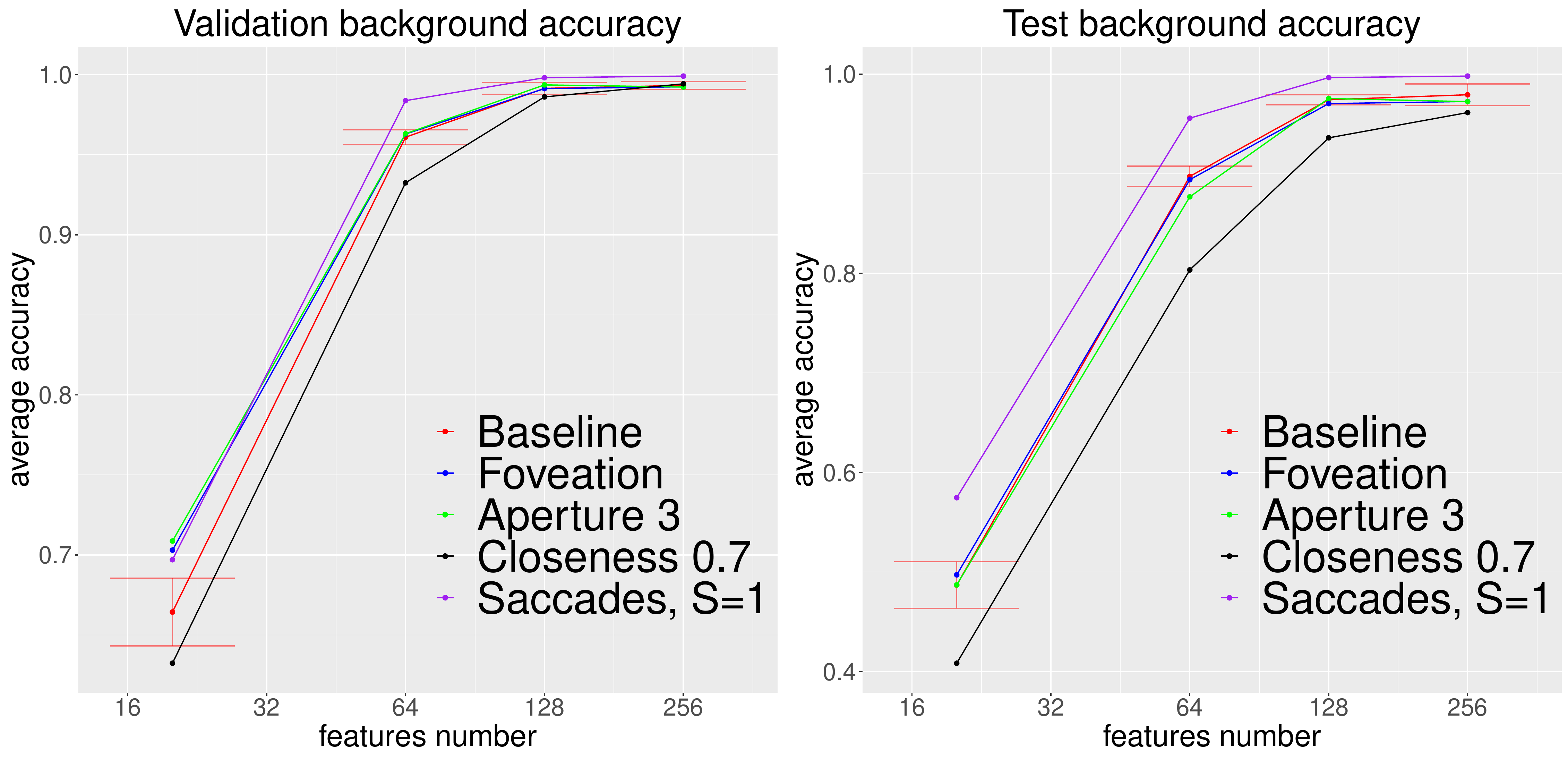}
    \caption{\textbf{Top:} Object recognition accuracy according to the applied visual statistics on the validation dataset (left) and test dataset (right). \textbf{Bottom:} Background recognition accuracy according to the applied visual statistics on the validation dataset (left) and test dataset (right). The baseline corresponds to perfectly clear images (No foveation and aperture of 20) of relatively distant objects.  The horizontal bars indicate the standard deviation of the baseline, we omit those of other curves for clarity.} 
    \label{fig:complex}
\end{figure}

\begin{figure}
    \centering
    \includegraphics[width=1\linewidth]{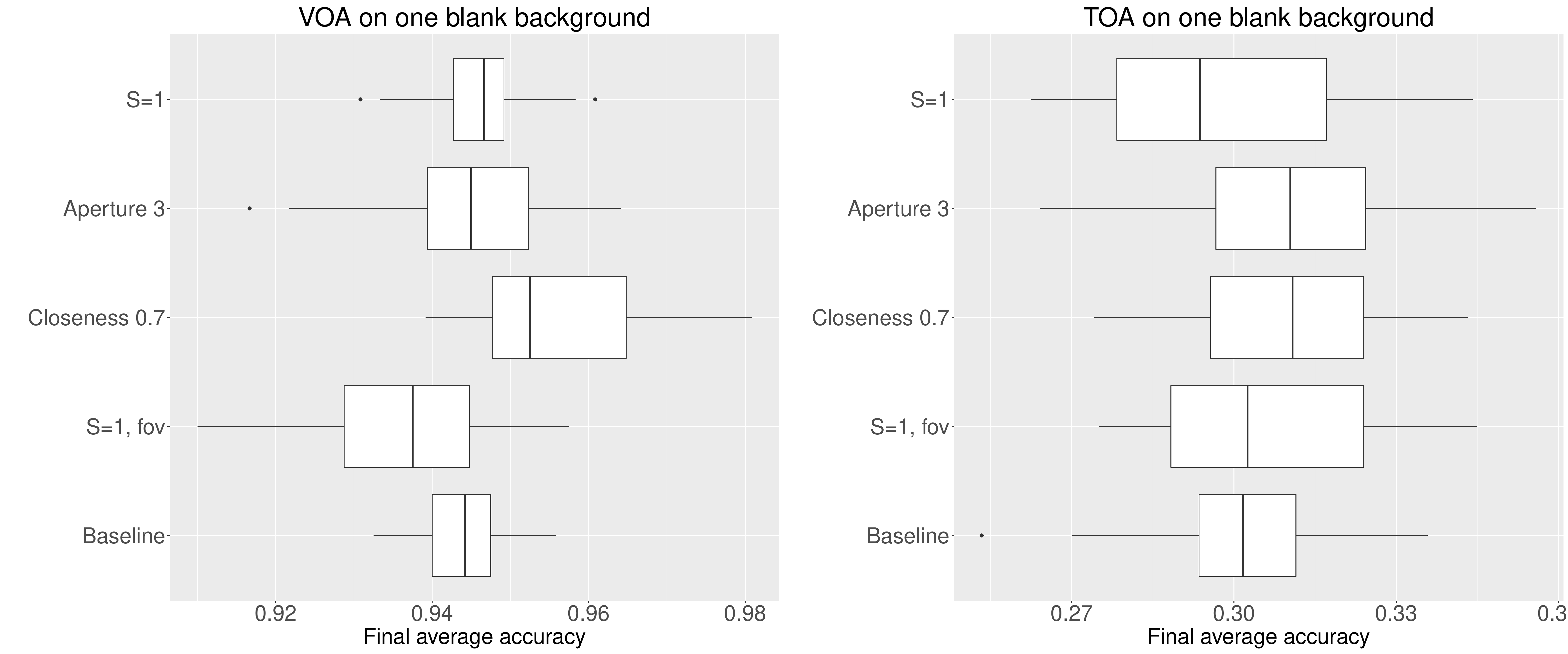}
    \caption{Final average object accuracy on the validation dataset (VOA, left) and the test dataset (TOA, right) with a representation trained on one blank background. We took the last 10 average accuracies (each 4000 steps between 360 000 and 400 000 timesteps) of the five seeds to compute the statistics. Since there is only one background, the agent does not turn its body but selects a new object every 10 timesteps to keep the same frequency of object switches as in other experiments.}
    \label{fig:whiteback}
\end{figure}
We train the representation of our agent with/without the visual effects of \secref{sec:method} on the environment displayed in \figref{fig:environments} and evaluate the representation on both validation and test datasets. In \figref{fig:complex}, we apply one transformation at a time. We first note that the effect of the visual statistics remain consistent over different sizes of latent space. 

\paragraph{Closer objects} \figref{fig:complex} (top) shows a steady increase of object recognition accuracy in both validation/test datasets with closer objects. We observe an opposite decrease in the background accuracy (bottom). Importantly, we observe in \figref{fig:whiteback} (left) better object representations (test dataset) even without different backgrounds; thus parts of the accuracy increase does not depend on background information. We deduce that part of the improvement comes from our CNN, presumably biased to better recognize larger objects. This is consistent with previous works \cite{bambach2018toddler} on supervised learning.

\paragraph{Low Saccade amplitudes} \figref{fig:complex} shows: 1) on the top-left graph, an increase of object recognition accuracy in both validation/test datasets with medium size of latent spaces (64 and 128); 2) on bottom graphs, a steady increase in the background recognition accuracy in validation/test datasets. As small saccades do not impact the representation when there is no background (see \figref{fig:whiteback}), this increased object recognition accuracy does not come from the reduced number of training/test diversity of examples (due to the lower amplitudes of spatial shifts). Instead, saccades generate higher shifts in backgrounds than on the objects, due to the higher distance of the background. Consequently, with high-amplitude saccades, a \textit{chair} visible in a background may be seen in one view, but not in another one. An agent needs to extract more than the \textit{chair} feature from the background to learn the similarity between views. Therefore, we conclude that a low amplitude of saccades reduces the number of features needed to learn saccade-invariant background representations, allowing the extraction of more object-oriented features. In compliance with our analysis in \secref{sec:backgrounds}, this effect vanishes with too-small or too-large latent spaces.

\begin{figure}[t]
    \centering
    \includegraphics[width=1\linewidth,height=3.5cm]{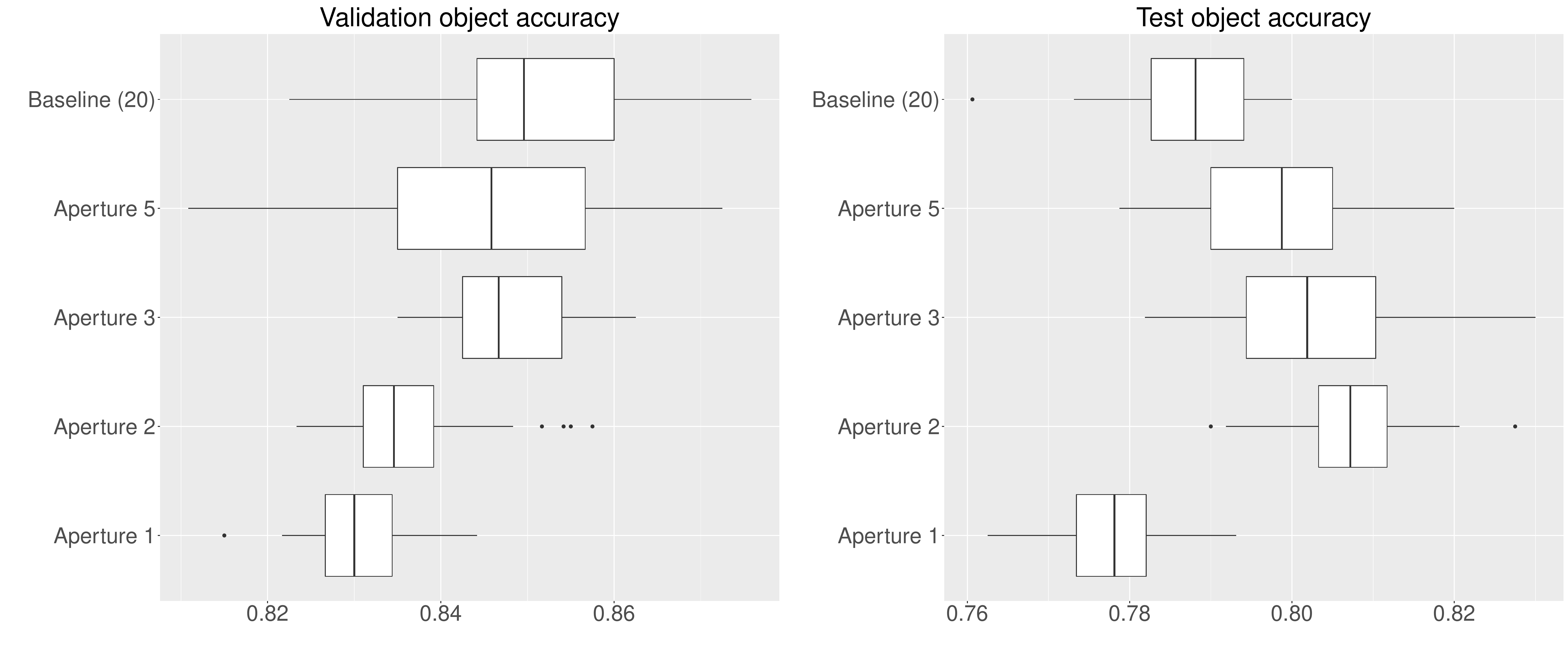}
    \includegraphics[width=1\linewidth, height=3.5cm]{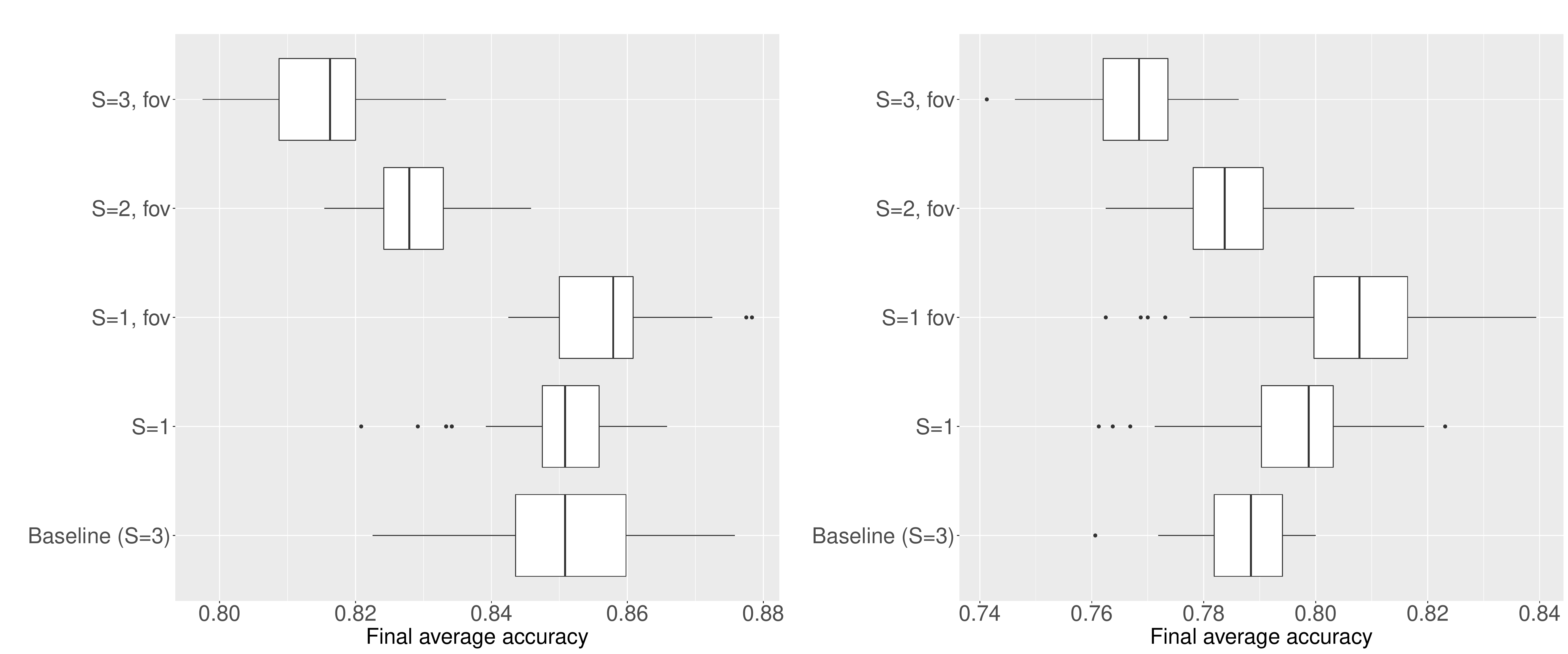}
    \caption{Object recognition accuracy on the validation dataset (left) and test dataset (right). We took the last 10 average accuracies (each 4000 steps between 360 000 and 400 000 timesteps) of the five seeds to compute the statistics. {Top:} Accuracy according to the number of aperture. Aperture 20 refers to the baseline. \textbf{Bottom}: Accuracy according to saccade amplitudes and the foveation of images.}
    \label{fig:apertures}
\end{figure}

\begin{figure}[t]
    \centering
    \includegraphics[width=1\linewidth]{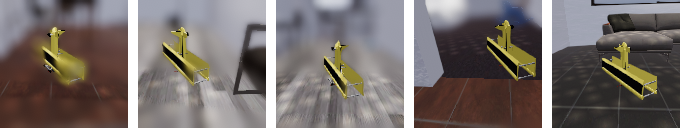}
    \caption{Illustration of different images of airplanes extracted from our validation dataset, with different saccade positions and backgrounds. From left to right, we use an aperture of 1, 2, 3, 5, 20.}
    \label{fig:airplanes}
\end{figure}

\paragraph{Limited depth of field} \figref{fig:complex} (top-right) exhibits a sound increase of object recognition accuracy with an aperture of $3$ only in the test datasets. To investigate its origin, we compare the impact of different aperture numbers on the object recognition accuracy in \figref{fig:apertures} (top). We observe a monotonic improvement on the validation object accuracy as we increase the aperture, saturating with an aperture of 3 and an important decrease with a low aperture (Aperture 1). This suggests that a too-low aperture, which blurs the object, prevents the extraction of important object features. We also observe a sweet spot for an aperture of 2 or 3 on the test object accuracy. The highest performances are thus obtained with a blurred background and slightly blurred parts of the object (see \figref{fig:airplanes}). To check whether the improvement comes from blurring small parts of the object or blurring the background, we trained the representation in the house environment using completely unblurred objects (Aperture 5 in \figref{fig:apertures}, top-right) and in front of a single white background (Aperture 3 in \figref{fig:whiteback}, right). A clear object in a blurred background already shows an improved test object recognition accuracy, and a blurred object in a clear background do not significantly impact the representation. Thus, we conclude that blurring the background is crucial to obtain this improved generalization. We hypothesize that a depth-wise blur increases the similarity of backgrounds between test and validation datasets, thereby increasing the generalization of objects on new backgrounds.

\begin{figure}
    \centering
    \includegraphics[width=1\linewidth, height=3.5cm]{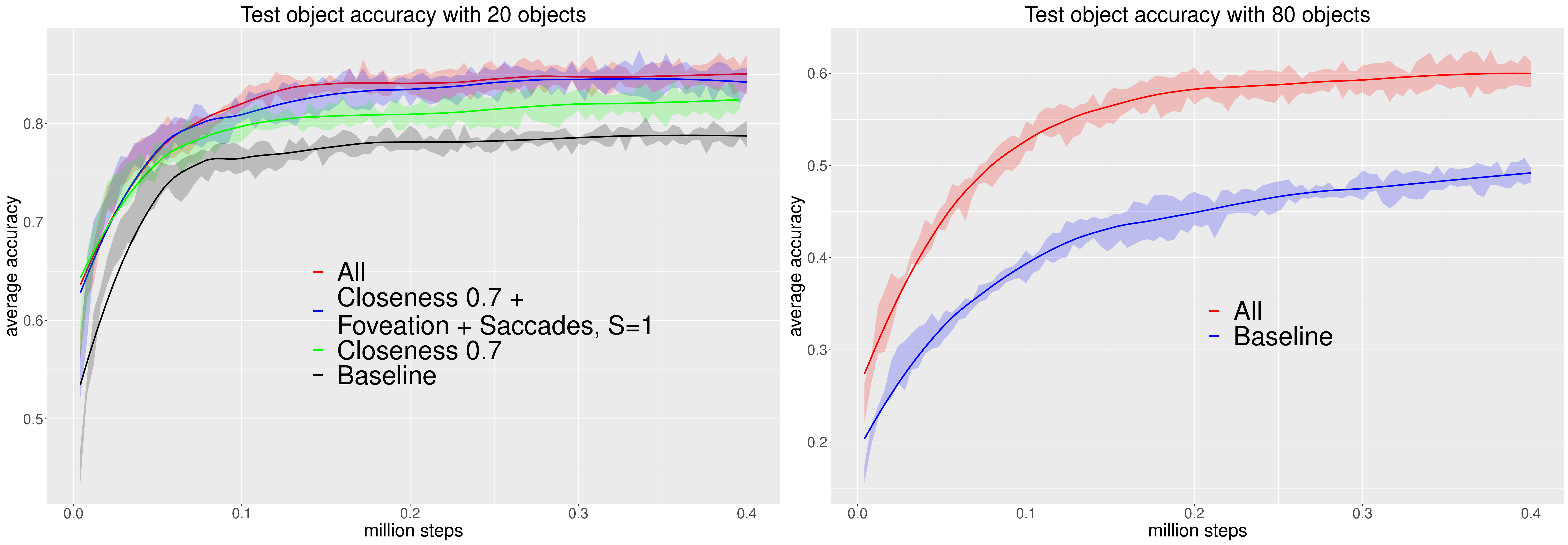}
    \caption{Test object recognition accuracy using toddler-inspired visual inputs on 20 (left) and 80 (right) objects. The \textit{All} experiment combines all image transformations. We set the aperture to $4$ instead of $3$ in the \textit{All} experiments to account for the increase of depth of field induced by closer objects in humans.}
    \label{fig:alls}
\end{figure}

\paragraph{Foveated images} \figref{fig:complex} (top) shows a that foveation can lead to worse object representation. To study the origin of this decrease, in \figref{fig:apertures} (bottom), we vary the range of amplitudes of saccades to control how much saccades direct the gaze towards the object or the background. A low amplitude makes, on average, saccades closer to the object and inversely. We observe that the foveation increases the object recognition accuracy only when the saccade amplitudes are small, such that most saccades end up on the object. Fixation of the object is thus crucial when considering foveated inputs. Similarly to the decrease of the depth of field, object-focused foveation mostly improves object recognition accuracy on the test dataset. As for the depth of field, we expect it comes from more similar features between test/validation backgrounds. This tends to be confirmed in \figref{fig:alls} (left), since the addition of aperture upon foveated images does not significantly improve the object recognition accuracy.

\paragraph{Combining all visual statistics} in \figref{fig:alls}, we progressively apply all visual statistics and assess the quality of the object representations.  We vary the number of objects to check whether our results scale with an increased object number. Overall, we clearly see that using toddler-inspired visual inputs improves object representations learnt through time-contrastive learning. 

\section{Conclusion}

We simulated viewing sequences as might be experienced by a toddler by letting an agent move and interact with different objects in different locations of a house. We hypothesized that time-contrastive learning, as an implementation of the slowness principle, models the unsupervised learning of visual representations in toddlers. First, we showed that the complexity and the number of backgrounds, negatively impact object representations learnt through time-contrastive learning in toddler-like environments. This suggests that object representations are improved by attenuating the information from the background. We then showed that incorporating viewing effects of a toddler aids the learning of object-oriented representations in a house environment. Our analysis suggests that foveation and a restricted depth of field have a similar effect of making object representations more robust to novel environments. We did not observe a consequent attenuation of background information. We also discovered that object-centered saccades are crucial to avoid the deterioration of the object representations due to 
foveation. Closer objects improve object recognition, presumably because they induce a greater bias towards object-centered features in neural networks \cite{bambach2018toddler}. However, while we did not show that closer objects attenuate the background information, we did not show either that this attenuation does not exist. Finally, using time-contrastive learning, a low amplitude of saccades seems to limit the extraction of features from the background. We showed that these three effects are cumulative. Overall, our contribution highlights the need to rethink the usual framework of contrastive learning as an embodied learning process.

As we found that low-amplitude saccades are the only visual condition that indeed slightly attenuates the extraction of background information, it remains a large margin of improvement in learning background-invariant representations. The solution may be to consider different frequencies of objects/backgrounds switches, since time-contrastive learning is expected to favour encoding slowly varying features. Furthermore, the incorporation of covert attention mechanisms seems promising.

Our study focuses on a small set of factors of variation in small environments, but other aspects also impact a toddler's visual input. For example, we do not consider other agents in the environment, the navigation of the agent between different rooms, cluttered scenes  or the visibility of arms/hands of the agent. Furthermore, we assume the agent executes normally distributed saccades and rotates objects independently from their view, which are inconsistent with findings from the developmental literature \cite{pereira2010early,trukenbrod2019spatial}. Playing is rather an active learning process where the toddler selects what he or she is interested in. As such, integrating intrinsic motivations \cite{aubret2019survey} to generate actions in an unsupervised way may be crucial for modeling the development of infants' object recognition skills.



\section*{Acknowledgment}
 This work was sponsored by a public grant overseen by the French National Agency through the IMobS3 Laboratory of Excellence (ANR-10-LABX-0016) and the IDEX-ISITE initiative CAP 20-25 (ANR-16-IDEX-0001). Financial support was also received from Clermont Auvergne Metropole through a French Tech-Clermont Auvergne professorship. 
We gratefully acknowledge support from the CNRS/IN2P3 Computing Center (Lyon - France) for providing computing and data-processing resources needed for this work. This work was also performed using HPC resources from GENCI–IDRIS (Grant 20XX-AD011011623R1). JT is supported by the Johanna Quandt Foundation. We thank Markus Ernst for his help with the foveation code.


\bibliography{references}
\bibliographystyle{IEEEtran}



\end{document}